\pdfoutput=1

\documentclass[11pt]{article}

\usepackage[final]{acl}

\usepackage{times}
\usepackage{latexsym}

\usepackage[T1]{fontenc}

\usepackage[utf8]{inputenc}

\usepackage{microtype}

\usepackage{inconsolata}

\usepackage{graphicx}

\usepackage{CJKutf8}
\usepackage{marvosym}
\title{ZZU-NLP at SIGHAN-2024 dimABSA Task: Aspect-Based Sentiment Analysis with Coarse-to-Fine In-context Learning}

\author{Senbin Zhu, Hanjie Zhao, Xingren Wang, Shanhong Liu, Yuxiang Jia\thanks{Corresponding author}, Hongying Zan \\
        School of Computer and Artificial Intelligence, Zhengzhou University, China\\
        \texttt{nlpbin@gs.zzu.edu.cn,
        \{hjzhao\_zzu,13257081272,18437919080\}@163.com}\\
        \texttt{ieyxjia@zzu.edu.cn}
    }

\begin{document}

\maketitle

\begin{abstract}
The DimABSA task requires fine-grained sentiment intensity prediction for restaurant reviews, including scores for Valence and Arousal dimensions for each Aspect Term. In this study, we propose a Coarse-to-Fine In-context Learning(CFICL) method based on the Baichuan2-7B model for the DimABSA task in the SIGHAN 2024 workshop. Our method improves prediction accuracy through a two-stage optimization process. In the first stage, we use fixed in-context examples and prompt templates to enhance the model's sentiment recognition capability and provide initial predictions for the test data. In the second stage, we encode the Opinion field using BERT and select the most similar training data as new in-context examples based on similarity. These examples include the Opinion field and its scores, as well as related opinion words and their average scores. By filtering for sentiment polarity, we ensure that the examples are consistent with the test data. Our method significantly improves prediction accuracy and consistency by effectively utilizing training data and optimizing in-context examples, as validated by experimental results.


\end{abstract}

\section{Introduction}
Aspect-Based Sentiment Analysis (ABSA) (Pontiki et al., 2014; 2015; 2016) is a critical NLP research topic that aims to identify the aspects of a given entity and analyze the sentiment polarity associated with each aspect. ABSA involves predicting tuples of sentiment elements for a given text, with four main elements constituting the focus of ABSA research: aspect term (a), aspect category (c), opinion term (o), and sentiment polarity (s)\citep{zhang2022survey}. 

Early studies on ABSA primarily focused on single sentiment elements such as aspect term \citep{liu2015fine, ma2019exploring}, aspect category \citep{zhou2015representation}, or sentiment polarity \citep{wang2016attention,chen2017recurrent}. However, recent research has introduced compound ABSA tasks involving multiple associated elements. These include Aspect Sentiment Triplet Extraction (ASTE) \citep{peng2020knowing,yuan2023encoding,chen2021bidirectional,mao2021joint,wu2020grid,xu2020position,zhang2020multi}, which extracts three elements in a triplet—aspect/target term, opinion term, and sentiment polarity.

Furthermore, Aspect Sentiment Quadruple Prediction (ASQP)\citep{zhang2021aspect,cai2021aspect,gao2022lego,mao2022seq2path,peper2022generative,zhou2023unified} extends ASTE by including an additional aspect category, thus constructing a quadruple. In contrast to representing affective states as discrete classes (i.e., polarity), there is also a dimensional approach that represents affective states as continuous numerical values, such as in the valence-arousal (VA) space \citep{russell1980circumplex}, providing more fine-grained emotional information \citep{lee2022chinese}. 
\begin{CJK}{UTF8}{gbsn}
For example, in the sentence “独家的鲔鱼抹酱超好吃。”, the corresponding elements are “鲔鱼抹酱” (aspect term), “食物\#品质” (aspect category), “超好吃” (opinion term), and “7.5\#7.25” (valence\#arousa score).
\end{CJK}

Resently, large language models (LLMs)\citep{brown2020language,touvron2023llama} have shown an impressive few-shot ability on several NLP tasks. To expect LLMs to perform better on few-shot tasks, in-context learning (ICL)\citep{dong2022survey} paradigm is becoming a flourishing research direction. This paradigm can generate a prediction of the test input by conditioning on few-shot input-output examples (also known as in-context examples or demonstrations), without requiring any updates to parameters. Previous studies \citep{liu2022makes,min2022rethinking} found that LLMs are highly sensitive to the choice of in-context examples. One typical strategy for retrieving helpful in-context examples is to leverage the overall semantic similarity between the candidate examples and test input. Further research has shown that retrieving highly relevant examples across multiple dimensions has achieved significant performance improvements in multi-domain ABSA tasks\citep{yang-etal-2024-faima-feature}.

Supervised Fine-Tuning (SFT) is a method that involves further training a pre-trained model using a labeled dataset to achieve better performance on specific tasks. In-context learning helps the model understand the task by providing a few examples during inference, but its performance is often limited by the selection and number of examples. SFT, on the other hand, directly trains on a large amount of labeled data, allowing the model to deeply understand various aspects of the task, thereby exhibiting higher accuracy and consistency in practical applications and achieving good performance on specific tasks\citep{zhang2024exploring}.

Our study addresses the DimABSA task at the SIGHAN 2024 workshop by proposing a two-stage context learning method based on the Baichuan2-7B\citep{yang2023baichuan} model to improve the accuracy of fine-grained sentiment intensity prediction for restaurant reviews. Our work consists of two main stages: In the first stage, we use fixed context examples to train the model, enhancing its ability to recognize sentiment elements. In the second stage, we utilize the Chinese BERT\citep{devlin2018bert} to encode the Opinion field and select the most similar training data as new context examples based on similarity calculations, thereby further improving the model's prediction accuracy and granularity. Experimental results show that our method significantly enhances the model's performance in both valence and arousal dimensions and effectively reduces sentiment polarity bias. Overall, our approach provides an efficient solution for the DimABSA task and offers valuable insights for the optimization of future fine-grained sentiment analysis models.



\section{Background}
The Chinese Dimensional Aspect-Based Sentiment Analysis (dimABSA)\citep{Lee2024Dimabsa} shared task is part of the SIGHAN 2024 workshop\footnote{https://dimabsa2024.github.io}. This task focuses on providing fine-grained sentiment intensity predictions for each extracted aspect of a restaurant review. The four sentiment elements are defined as follows:

Aspect Term (A): Denotes an entity indicating the opinion target. If the aspect is omitted and not mentioned clearly, "NULL" is used to represent the term.
Aspect Category (C): Represents a predefined category for the explicit aspect within the restaurant domain. The categories are based on the SemEval-2016 Restaurant dataset \citep{pontiki2016semeval} and include twelve categories, each split into an entity and attribute using the symbol “\#”. 
Opinion Term (O): Describes the sentiment words or phrases related to the aspects.
Sentiment Intensity (I): Reflects the sentiments using continuous real-valued scores in the valence-arousal dimensions. Valence indicates the degree of pleasantness (positive or negative feelings), while arousal indicates the degree of excitement or calmness. Both dimensions use a nine-degree scale, where 1 denotes extremely high-negative or low-arousal sentiment, 9 denotes extremely high-positive or high-arousal sentiment, and 5 denotes neutral or medium-arousal sentiment. 
This task aims to evaluate the capability of automatic systems for Chinese dimensional ABSA and is divided into three subtasks:

Subtask 1: Intensity Prediction: Focuses on predicting sentiment intensities in the valence-arousal dimensions. Given a sentence and a specific aspect, the system should predict the valence-arousal ratings.
Subtask 2: Triplet Extraction: Aims to extract sentiment triplets composed of three elements (aspect, opinion, intensity) from a given sentence.
Subtask 3: Quadruple Extraction: Aims to extract sentiment quadruples composed of four elements (aspect, category, opinion, intensity) from a given sentence.

Our team chose to participate in the more challenging second and third subtasks, and we achieved third place in the evaluation task.

\section{System Overview}

\begin{figure*}[h]
\centering 
  \includegraphics[width=1.0\linewidth]{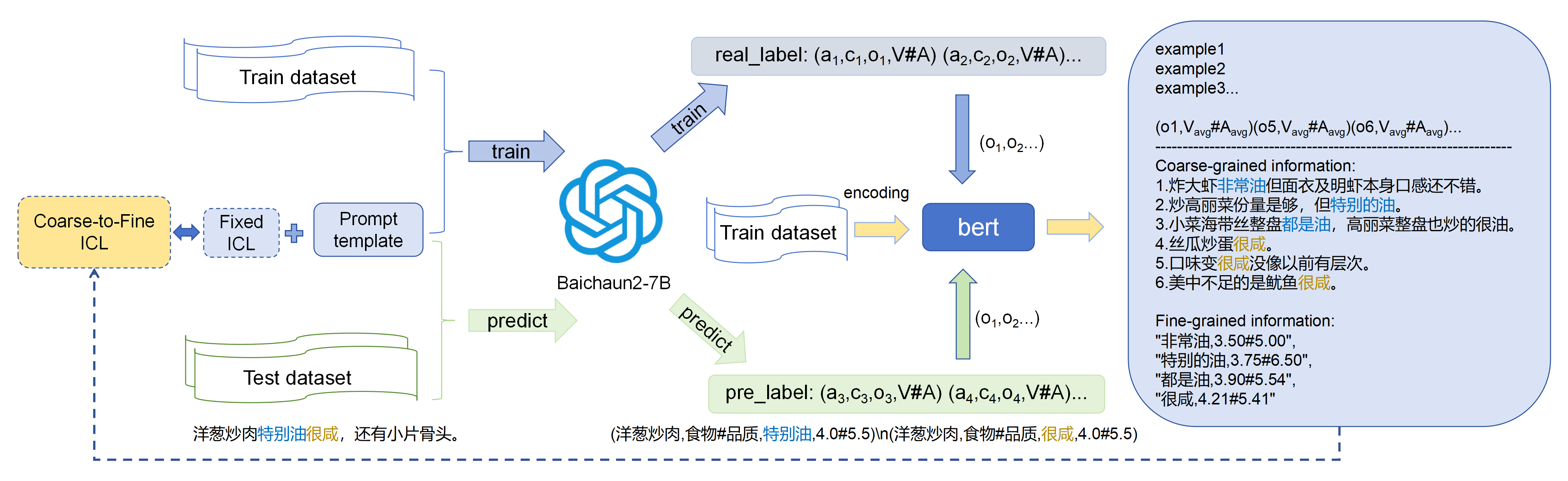}
  \caption {The architecture of our system. The figure illustrates a two-stage in-context learning method based on the Baichuan2-7B model to improve prediction accuracy in the DimABSA task. In the first stage, fixed in-context examples (Fixed ICL) are used to process training data. The model's sentiment recognition ability is enhanced through a prompt template, and initial predictions are made for the test data. In the second stage, the Opinion field is encoded using BERT, and the most similar training data is selected as new in-context examples based on similarity. These examples include the Opinion field and its scores, as well as related opinion words and average scores. Sentiment polarity filtering ensures that the in-context examples are consistent with the test data. Finally, these new in-context informations are input into the model along with the test data for re-prediction, yielding optimized quadruple results.}
  \label{fig:CFCL}
\end{figure*}


We use Baichuan2-7B as the base model and propose a two-stage context learning method to improve prediction accuracy in the DimABSA task. This method incrementally optimizes the model output through preliminary and refined prediction stages, fully utilizing the information in the training data. Our framework is shown in Figure~\ref{fig:CFCL}.

Baichuan2\citep{yang2023baichuan} is a Chinese and English bilingual language model. It achieved the best performance among models of the same size on standard benchmarks(C-Eval\citep{huang2024c}, MMLU\citep{hendrycks2020measuring}).

\subsection{Fixed In-context Learning Stage}

In the first stage, we utilize a few-shot learning method to process the training data. Specifically, we prepare three fixed context examples for each training sample and input these examples along with the training data into the model. This approach allows the model to learn task-related features from limited context information. After training, we use the trained model to predict the test data and obtain preliminary quadruplet results (aspect, category, opinion, intensity).

\subsection{Example Retrieval Enhancement Stage}
The objective of the second stage is to further enhance the model's prediction accuracy through similarity calculation and context example optimization. First, we use a BERT model to encode the Opinion field of each data label and calculate the cosine similarity between the Opinion encoding of each test data and that of each training data. The similarity calculation results are used to select the three most similar training data as new context examples. These examples include not only the Opinion fields and their scores but also related opinion words and their average scores, providing more detailed reference information.

To prevent significant bias in emotional polarity, we filter candidate examples based on the Valence scores predicted in the first stage, ensuring that the selected context examples are consistent with or similar to the current test data in terms of emotional polarity. Then, we input the new context examples along with the test data into the model for re-prediction, ultimately obtaining the optimized quadruplet prediction results.

This two-stage method significantly improves the model's prediction performance. The preliminary prediction in the first stage lays the foundation for the refined prediction in the second stage. The second stage, through similarity calculation and context example optimization, further enhances the accuracy and granularity of the prediction results. The overall method not only fully utilizes the information in the training data but also effectively reduces the impact of emotional polarity bias through careful filtering and context construction.



\section{Experimental Setup}

\subsection{Dataset}
The DimABSA dataset provided by the evaluation organizers includes 6000 training samples, 100 validation samples, and 2000 test samples of restaurant reviews. These data provide a substantial foundation for model training, validation, and final evaluation. The innovation of this evaluation task lies in the requirement to assign scores for valence and arousal dimensions to each aspect term, which is also the main challenge of the task.
  

\begin{figure*}[h]
  \includegraphics[width=0.48\linewidth]{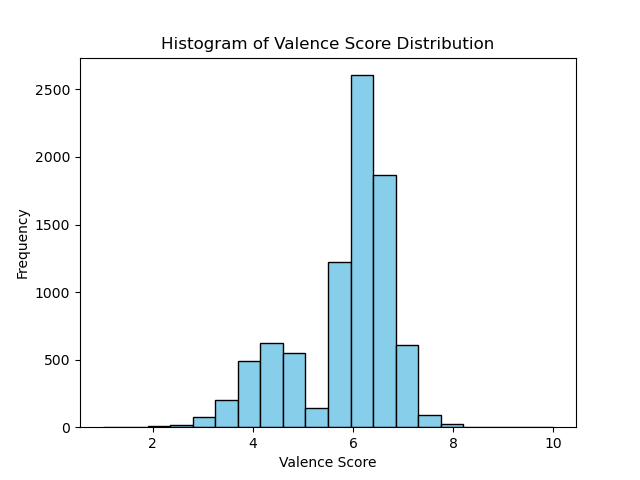} \hfill
  \includegraphics[width=0.48\linewidth]{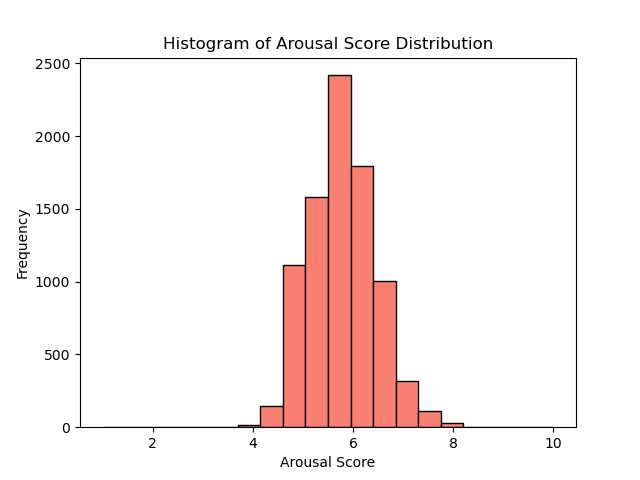}
  \caption {The distribution of valence and arousal scores of train dataset}
    \label{fig:VandA}
\end{figure*}

\begin{figure}[h]
  \includegraphics[width=\columnwidth]{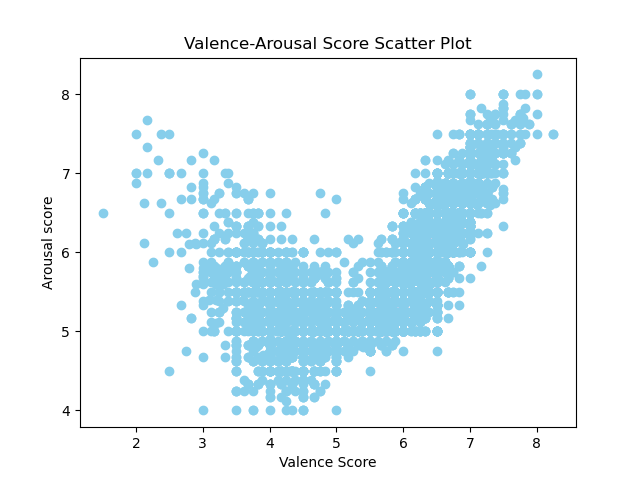}
  \caption{The distribution of continuous real-valued scores in the valence-arousal dimensions}
  \label{fig:VA}
\end{figure}

We conduct a detailed examination of the sample distribution in these two dimensions and the correlation between the scores. Figure~\ref{fig:VandA} shows the sample distribution for the valence and arousal dimensions, respectively. It can be seen that there are more samples with low valence scores (4-5) and more samples with high valence scores (6-7). The majority of arousal scores are concentrated between 5 and 7, with fewer samples at extreme values. Figure~\ref{fig:VA} shows the scatter plot of valence and arousal scores, illustrating the relationship between these two dimensions.

Through our analysis, we find that the distribution characteristics of this dataset align with those of The Chinese EmoBank~\citep{lee2022chinese}, a dimensional sentiment resource. The reasonable distribution of valence and arousal dimensions provides authentic and effective data support for model training, helping the model to make accurate predictions under different levels of emotional intensity.




\subsection{Implementation Details}
We use Baichuan2-7B as our base model. During training, we use a batch size of 4 and a gradient accumulation step size of 4. We further employ the Adam optimizer with a learning rate of $8 \times 10^{-5}$. The training employs the LoRA efficient tuning method with precision set to fp16. We conduct the training on an NVIDIA V100 GPU.

\subsection{Evaluation Metrics}

First, the valence and arousal values are rounded to an integer. Next, a triplet/quadruple is regarded as correct if and only if the three/four elements and their combination match those in the gold triplet/quadruple. On this basis, we calculate the Precision, Recall, and F1-score as the evaluation metrics, defined as the following equations.

\begin{equation}
  \label{eq:example}
  Precision = \frac{TP}{TP+FP}
\end{equation}

\begin{equation}
  \label{eq:example}
  Recall = \frac{TP}{TP+FN}
\end{equation}

\begin{equation}
  \label{eq:example}
  F1 = \frac{2*Precision*Recall}{Precision+Recall}
\end{equation}
where TP, FP, and FN denote true positives, false positives, and false negatives, respectively. Precision is defined as the percentage of triplets/quadruples extracted by the system that are correct. Recall is the percentage of triplets/quadruples present in the test set found by the system. The F1-score is the harmonic mean of precision and recall. All metrics range from 0 to 1. A higher Precision, Recall, and F1 score indicate more accurate performance. A system’s overall ranking is based on the F1 score. The higher the F1 score, the better the system performance.
\begin{table*}[t]
  \centering
  \begin{tabular}{llll}
    \hline
    \textbf{method} & \textbf{V-Quat-F1} & \textbf{A-Quat-F1} & \textbf{VA-Quat-F1} \\
    \hline
    \verb|4 shot+Prompt1|       & {0.53}    & {0.38}    & {0.27}   \\
    \verb|4 shot+Prompt2|       & {0.54}    & {0.46}    & {0.32}   \\
    \verb|4 shot+Instruction Tuning|     & {0.61}     & {0.47}    & {0.32}   \\
    \verb|Coarse-to-Fine ICL+Instruction Tuning|      & {0.62}    & {0.51}    & {0.38}   \\\hline
  \end{tabular}
  \caption{\label{eval result}
    Evalution dataset results for quadruple extraction task
  }
\end{table*}

\begin{table*}[h]
  \centering
  \begin{tabular}{llll}
    \hline
    \textbf{task} & \textbf{V-F1} & \textbf{A-F1} & \textbf{VA-F1} \\
    \hline
    \verb|Triplet Extraction|    & {0.542}    & {0.507}    & {0.389}\\
    \verb|Quadruple Extraction|    & {0.522}    & {0.489}    & {0.376}   \\\hline
  \end{tabular}
  \caption{\label{test result}
    The result of test dataset for triplet and quadruple extraction
  }
\end{table*}
\subsection{Evaluation Results}
Each metric for the valence and arousal dimensions is calculated and ranked either independently or in combination. Precision is defined as the percentage of triplets/quadruples extracted by the system that are correct. Recall is the percentage of triplets/quadruples present in the test set found by the system. The F1-score is the harmonic mean of precision and recall. All metrics range from 0 to 1. A higher Precision, Recall, and F1 score indicate more accurate performance.

Previous research has shown that within certain limits, the performance of large language models improves with an increasing number of context examples. Considering computational constraints, we fine-tune Baichuan2-7B using four manually selected context examples. The selection criteria aim to ensure that examples are diverse and representative of the most common features in the dataset, thereby optimizing model performance to the greatest extent.

Additionally, adjustments to the prompt template significantly impact model performance. We use ChatGPT to optimize the logic and content quality of the prompt templates, emphasizing specific task characteristics and common pitfalls to further refine the templates. The initial template (prompt1) and the optimized template (prompt2) will be shown in the appendix. Experimental results indicate that the optimized prompt2 improves performance by five percentage points.

Instruction-tuning is a method to enhance the model's understanding of task instructions, thereby improving its generalization ability in specific tasks. Based on the above strategy for prompt adjustment, we design ten task templates for the model to randomly choose from, aiming to help the model comprehensively understand the task. After incorporating instruction-tuning, the V-Quat-F1 and A-Quat-F1 scores of the model's predictions improve, but the VA-Quat-F1 score shows no significant change. This suggests that while the model's understanding of valence and arousal dimensions improves individually, it does not adequately address the consistency between these two dimensions.

Given the challenge of this task, which requires scoring aspect sentiment in two dimensions—particularly the more difficult arousal dimension—we further optimize context examples using initially predicted test set label information and provide the model with more granular word-level standard score information. Specifically, in the second stage, we use an example retrieval method to find 3 to 5 context examples for each data sample and provide more than three word-level standard score examples. By providing dual-granularity information (sentence-level context examples and word-level standard scores), the prediction scores for both valence and arousal dimensions improve further. More importantly, the consistency between these two dimensions also significantly improves, with the VA-Quat-F1 score reaching 0.38. Detailed experimental results on the validation set are shown in Table~\ref{eval result}.

Finally, the test results on the test set are shown in Table~\ref{test result}. For the triplet extraction task, we ignore the "category" aspect and adopt the same strategy, achieving good results as well. This further validates the effectiveness and broad applicability of our method.






\subsection{Case Study}
\begin{figure*}[h]
\centering 
  \includegraphics[width=1.0\linewidth]{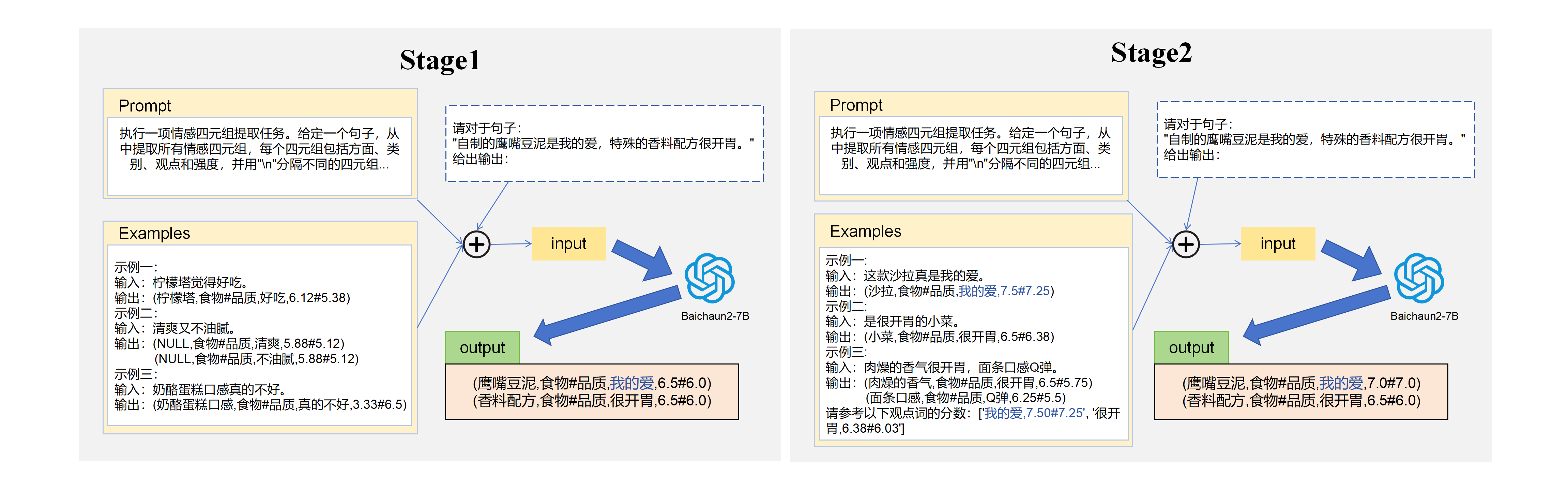}
  \caption {Case Study}
  \label{fig:case}
\end{figure*}
As shown in figure~\ref{fig:case}, after adopting the optimized examples, the reference for the term "my love" in the predictions becomes more precise. Initially, when using fixed examples, the valence and arousal scores for "my love" are 6.5 and 6.0, respectively. Although these scores reflect a certain level of emotional intensity, they are not entirely accurate. By employing optimized examples in the second phase, we provide more relevant context. In the optimized example, the sentence "This salad is really my love." corresponds to valence and arousal scores of 7.50 and 7.25, respectively. These scores capture the emotional nuances more effectively. Considering this more fitting example, we re-predict the scores for "Homemade hummus is my love," resulting in adjusted valence and arousal scores of 7.0 and 7.0. These revised scores are more reasonable, demonstrating that using optimized context examples can significantly improve the accuracy and consistency of predictions, thereby better meeting the demands of fine-grained sentiment analysis.

\section{Conclusion}
This study proposes a two-stage context learning method based on the Baichuan2-7B model for the DimABSA task at the SIGHAN 2024 workshop. The task requires fine-grained sentiment intensity prediction for restaurant reviews.

In the first stage, we enhance the model's sentiment element recognition ability using fixed context examples. In the second stage, we utilize BERT to encode the Opinion field and select the most similar training data based on similarity calculation as new context examples. These relevant examples improve the accuracy of sentiment intensity prediction.

Experimental results show that our two-stage method significantly enhances the accuracy and granularity of predictions. The method effectively utilizes training data and reduces sentiment polarity bias.

Overall, our approach provides an efficient solution for the DimABSA task and offers valuable insights for optimizing future models for fine-grained sentiment analysis.





\section*{Limitations}
Although our proposed method demonstrates significant performance improvements in the DimABSA task, there are still some limitations. First, our method focuses on enhancing the accuracy of sentiment intensity prediction, without further optimization for the Aspect field and its Category. Second, the success of the second stage is relatively dependent on the accuracy of the similarity measure between the Opinion field and the training data, and the issue of error propagation requires further analysis and discussion. Additionally, our method has high computational resource demands, especially when performing large-scale data training and optimization. This could limit its practicality and widespread adoption in real-world applications.

\section*{Acknowledgments}
We express our gratitude to the organizers for providing such an inspiring competition and resolving our questions patiently. We would like to express our sincere gratitude to those who have provided detailed suggestions for our paper. Their diligent efforts are greatly appreciated.
\bibliography{custom}

\appendix

\section{Appendix}
\label{sec:appendix}
Prompt 1:
\begin{CJK}{UTF8}{gbsn}
执行一项情感四元组提取任务。给定一个句子，提取句子中所有的情感四元组(方面、类别、观点、强度)，用"\textbackslash n"分隔不同的四元组。

每个四元组中的“方面”是句子中被评价对象的特定方面或特征，如果省略了该方面而没有明确提及，使用“NULL”来表示该术语。
”类别“是预定义的12种类别之一，根据常识判断。预定义类别：餐厅\#概括、餐厅\#价格、餐厅\#杂项、食物\#价格、食物\#品质、食物\#份量与款式、饮料\#价格、饮料\#品质、饮料\#份量与款式、氛围\#概括、服务\#概括、地点\#概括。
”观点“是对方面的情感词或短语。
”强度“指效价-唤醒的二维情绪强度，其中效价代表情绪体验的整体愉悦程度(高兴-不高兴)，唤醒代表情绪的强度水平(平静-兴奋)，每个指标的范围应为 1.0 到 9.0。在效价和唤醒维度上的值为 1 表示极度负面和低唤醒情感，相反，9 表示极度正面和高唤醒情感，5 表示中性和中等唤醒情感。精确到小数点后两位，效价-唤醒值以\#分隔。

示例如下：
{examples}

请对于句子：

给出输出：
\end{CJK}

Prompt 1 in English:
Extract sentiment quads. Given a sentence, extract all sentiment quads (aspect, category, opinion, intensity) in the sentence, separated by "\textbackslash n".

In each quad, the "aspect" is the specific aspect or feature of the object being evaluated in the sentence. If the aspect is omitted or not explicitly mentioned, use "NULL" to represent the term.
The "category" is one of the predefined 12 categories, determined based on common sense. The predefined categories are: Restaurant\#General, Restaurant\#Price, Restaurant\#Miscellaneous, Food\#Price, Food\#Quality, Food\#Portion and Style, Drink\#Price, Drink\#Quality, Drink\#Portion and Style, Ambience\#General, Service\#General, Location\#General.
The "opinion" is the sentiment word or phrase describing the aspect.
The "intensity" refers to the two-dimensional emotion intensity of Valence-Arousal, where valence represents the overall pleasantness of the emotional experience (happy-unhappy), and arousal represents the intensity level of the emotion (calm-excited). Each indicator ranges from 1.0 to 9.0. A value of 1 on the valence and arousal dimensions indicates extremely negative and low arousal emotions, respectively, while 9 indicates extremely positive and high arousal emotions, and 5 indicates neutral and medium arousal emotions. Values should be precise to two decimal places, with valence and arousal values separated by \#.

Example:
{examples}

For the sentence:

Provide the output:\\

Prompt 2:
\begin{CJK}{UTF8}{gbsn}
执行一个情感四元组提取任务。给定一个句子，从中提取所有情感四元组，其中包括方面、类别、观点和强度，并用"\textbackslash n"分隔不同的四元组。

每个四元组包括以下要素：
“方面”指的是句子中被评价对象的具体方面或特征。如果没有明确提及方面，则使用“NULL”表示。
“类别”是根据常识判断的预定义类别之一，共有12种。预定义类别包括：餐厅\#概括、餐厅\#价格、餐厅\#杂项、食物\#价格、食物\#品质、食物\#份量与款式、饮料\#价格、饮料\#品质、饮料\#份量与款式、氛围\#概括、服务\#概括、地点\#概括。
“观点”是对被评价对象特定方面的情感词或短语。
“强度”表示情感的效价和唤醒，分别代表情绪体验的整体愉悦程度（高兴-不高兴）和情绪的强度水平（平静-兴奋）。效价和唤醒的范围是1.0到9.0，其中1表示极度负面和低唤醒情感，9表示极度正面和高唤醒情感，5表示中性和中等唤醒情感。效价和唤醒值以\#分隔，精确到小数点后两位。
输出格式应严格按照以下示例的格式：
(方面, 类别, 观点, 强度)
每个四元组在括号内，不要输出无关信息。
观点是最细粒度的情感词，需要为每一个提取出的观点生成相应的四元组。

示例如下：
{examples}

请对于句子：

给出输出：
\end{CJK}

Prompt 2 in English:
Execute a task of extracting sentiment quadruples. Given a sentence, extract all sentiment quadruples from it, including aspect, category, opinion, and intensity, and separate different quadruples with "\textbackslash n".

Each quadruple includes the following elements:

"Aspect" refers to the specific aspect or feature of the evaluated object in the sentence. If the aspect is not explicitly mentioned, use "NULL" to represent it.
"Category" is one of the predefined categories judged based on common sense. There are 12 predefined categories: Restaurant\#General, Restaurant\#Prices, Restaurant\#Miscellaneous, Food\#Prices, Food\#Quality, Food\#Style and Options, Drinks\#Prices, Drinks\#Quality, Drinks\#Style and Options, Ambience\#General, Service\#General, and Location\#General.
"Opinion" is the emotional word or phrase regarding the specific aspect of the evaluated object.
"Intensity" represents the valence and arousal of the emotion, where valence indicates the overall pleasantness of the emotional experience (happy-unhappy) and arousal indicates the intensity level of the emotion (calm-excited). The range of valence and arousal is from 1.0 to 9.0, where 1 indicates extremely negative and low arousal emotion, 9 indicates extremely positive and high arousal emotion, and 5 indicates neutral and moderate arousal emotion. The valence and arousal values are separated by \# and precise to two decimal places.
The output format should strictly follow the format of the following example:
(aspect, category, opinion, intensity). 
Each quadruple should be enclosed in parentheses, and do not output any irrelevant information. Each opinion is the most granular emotional word, and a corresponding quadruple should be generated for each extracted opinion.

Example:
{examples}

Given the sentence:

Provide the output:

\end{document}